\documentclass{article}
\usepackage{arxiv}
\usepackage{hyperref}       % hyperlinks
\usepackage{url}            % simple URL typesetting
\usepackage{booktabs}       % professional-quality tables
\usepackage{amsfonts}       % blackboard math symbols
\usepackage{nicefrac}       % compact symbols for 1/2, etc.
\usepackage{microtype}      % microtypography
\usepackage{lipsum}		% Can be removed after putting your text content
\usepackage{graphicx}
\usepackage{natbib}
\usepackage{doi}

\title{Smart-Hiring: An Explainable end-to-end Pipeline for CV Information Extraction and Job Matching}

\author{ \href{https://orcid.org/0000-0002-3794-844X}{\hspace{1mm}Kenza KHELKHAL, Dihia LANASRI}\\
	ATM Mobilis\\
	Algiers, Algeria\\	
	\texttt{khelkhalkenza88@gmail.com , ad\_lanasri@esi.dz} \\
    \\    
}

\renewcommand{\shorttitle}{Smart Hiring Pipeline}

\hypersetup{
pdftitle={Smart Hiring: Resume Parsing and Job Matching},
pdfsubject={NLP, ANALYTICS},
pdfauthor={K.KHELKHAL and al},
pdfkeywords={Natural Language Processing, Resume Parsing, Job Matching, Explainable AI},
}

\begin{document}
\maketitle

\begin{abstract}  %%ok%%%
Hiring processes often involve the manual screening of hundreds of resumes for each job, a task that is time and effort consuming, error-prone, and subject to human bias.

This paper presents Smart-Hiring, an end-to-end Natural Language Processing (NLP) pipeline designed to automatically extract structured information from unstructured resumes and to semantically match candidates with job descriptions.

The proposed system combines document parsing, named-entity recognition, and contextual text embedding techniques to capture skills, experience, and qualifications.

Using advanced NLP technics, Smart-Hiring encodes both resumes and job descriptions in a shared vector space to compute similarity scores between candidates and job postings.
The pipeline is modular and explainable, allowing users to inspect extracted entities and matching rationales.

Experiments were conducted on a real-world dataset of resumes and job descriptions spanning multiple professional domains, demonstrating the robustness and feasibility of the proposed approach.

The system achieves competitive matching accuracy while preserving a high degree of interpretability and transparency in its decision process.
This work introduces a scalable and practical NLP framework for recruitment analytics and outlines promising directions for bias mitigation, fairness-aware modeling, and large-scale deployment of data-driven hiring solutions.
\end{abstract}


\section{Introduction} %%ok%%%
Recruitment is a critical yet time-consuming process in modern organizations, often requiring human resources professionals to manually review hundreds of resumes for each job opening. This manual screening not only demands significant effort and time but also introduces inconsistencies and biases in candidate selection. The increasing availability of digital resumes and online job platforms has created an urgent need for automated, intelligent systems capable of efficiently analyzing candidate profiles while maintaining fairness, transparency, and interpretability in the decision-making process.

Traditional approaches used for automatic or semi-automatic recruitment typically rely on keyword or rule-based matching, which are limited in their ability to capture the semantic relationships between job descriptions and candidate qualifications. For example, systems that depend on exact keyword matches may fail to recognize that “software developer” and “application engineer” represent semantically similar roles. 
Moreover, extracting information from unstructured resumes represent a challenging task. Most CVs are not respecting the ATS standard, identifying the right information in the correct position in each CV still a complex problem to resolve. Recent advances in Natural Language Processing (NLP) and deep learning—particularly with transformer-based language models—offer new opportunities for overcoming these limitations through context-aware understanding of textual information.

To overcome these limits, we present \textbf{Smart-Hiring}, an end-to-end NLP pipeline designed to automate two fundamental stages of the recruitment process: (1) the extraction of structured information from heterogeneous and unstructured resumes, and (2) the semantic matching between candidate profiles and job descriptions. The system combines rule-based heuristics, lightweight machine learning models, and contextual text embeddings to achieve robust information extraction and semantically meaningful candidate-job alignment.

Unlike conventional recruitment automation systems, Smart-Hiring emphasizes \textit{explainability} and \textit{trustworthiness} as core design principles. In addition to providing similarity scores between candidates and job postings, the system highlights the most influential factors contributing to each recommendation—such as matching skills, experience, or education—thereby enabling recruiters to understand and audit the reasoning behind automated decisions.

To evaluate the feasibility and effectiveness of the proposed framework, we conducted experiments on a real-world dataset containing more than a thousand of resumes and hundreds of job descriptions spanning multiple professional domains. The results demonstrate that Smart-Hiring achieves competitive accuracy while maintaining interpretability, confirming its potential to significantly reduce manual workload during pre-screening and enhance transparency in AI-assisted hiring.

This paper is organized as follows: a \textbf{\textit{Related Work}} Section which gives an overview of main works published in CV-JD matching domain. Section \textbf{\textit{Resume Information Extraction}} describes the resume parsing and information extraction process. Section \textbf{\textit{Job Matching}} details the job matching methodology. Section \textbf{\textit{Experiments and Results Discussion}}, presents experimental results and analysis. Finally, the paper concludes with a discussion of limitations, potential improvements, and directions for future research.

\section{Related Work} %%ok%%%
Automated recruitment and candidate-job matching have received growing attention in recent years, driven by the availability of online professional data and advances in Natural Language Processing (NLP). 

Research on automated recruitment systems spans two primary areas: (1) \textit{resume information extraction} and (2) \textit{job–candidate matching}. Both components are critical for building intelligent and interpretable AI-driven hiring solutions.

\subsection{Resume Information Extraction}
Information extraction from resumes aims to transform heterogeneous, semi-structured documents into structured representations suitable for downstream analytics. Early systems relied on handcrafted regular expressions and rule-based templates to detect entities such as names, skills, and education \cite{jain2018automated, chen2019resume}. While such systems performed reasonably well on uniform layouts, they struggled with visual variability, nonstandard section headers, and multilingual resumes.

To overcome these limitations, recent studies have applied machine learning and deep learning to model the complex structures of resumes. \cite{ma2018resumeparser} employed a Conditional Random Field (CRF) model to label textual segments into predefined categories, achieving high precision for contact and education fields. \cite{li2020deep} introduced a hybrid CNN–BiLSTM–CRF architecture for named entity recognition (NER) on resumes, significantly improving generalization to unseen layouts. Other works, such as \cite{xu2021layoutlm}, leveraged layout-aware document understanding models like LayoutLM, which jointly encode textual and spatial information from PDFs to enhance extraction accuracy on visually rich documents.

Publicly available datasets for this task remain limited due to privacy concerns, but notable examples include the \textit{Resume Entities Dataset} (Kaggle, 2019), the \textit{RESUME-NER Corpus} \citep{li2020deep}, and the \textit{RVL-CDIP dataset} for layout-based document analysis \citep{harley2015icdar}. These resources have facilitated the development of benchmarks and comparative studies on resume parsing performance.

\subsection{Job–Candidate Matching}
Candidate-job matching focuses on measuring the semantic compatibility between candidate profiles and job descriptions. Traditional approaches relied on keyword or vector space similarity measures \citep{malinowski2006matching, zhao2015application}, which failed to capture contextual relationships between terms. The advent of distributed word representations such as Word2Vec and GloVe \citep{mikolov2013efficient, pennington2014glove} introduced semantic matching, later enhanced by contextual embeddings like BERT and Sentence-BERT \citep{devlin2019bert, reimers2019sentence}.

\citet{ghodsi2020automated} proposed an embedding-based model to compute cosine similarity between resume and job description vectors, improving ranking quality. \citet{wong2021bert} demonstrated that fine-tuning BERT on recruitment corpora yields superior performance in candidate ranking tasks. More recently, \citet{chen2022semantic} introduced a transformer-based ranking model incorporating attention over skills and experience, demonstrating interpretable matching performance.

\subsection{Discussion}
Despite substantial progress in both subfields, most existing systems treat resume parsing and job matching as independent processes, resulting in information loss between stages. Furthermore, explainability and bias mitigation remain underexplored in commercial recruitment AI. \textbf{\textit{Smart-Hiring}} addresses these gaps by integrating robust information extraction, semantic matching, and interpretability within a single, unified pipeline for end-to-end recruitment analytics.

\section{Resume Information Extraction}
The first stage of the \textbf{\textit{Smart-Hiring}} pipeline focuses on transforming unstructured and heterogeneous resumes into structured, useful data. This task is particularly challenging since candidates use a wide variety of CV layouts, often containing multiple columns, icons, or tabular elements. Furthermore, resumes may be written in different languages, such as French and English, and rely on non-standard section titles (e.g., Formation, Education, Parcours Professionnel), which complicates automatic segmentation and understanding. Hence, a robust and adaptable extraction process is required to ensure consistent information retrieval across diverse formats.

The extraction process begins with the conversion of resume files into raw text. For standard text-based PDFs, the system employs \texttt{pdfplumber}, a lightweight library designed for text and layout extraction. The tool not only retrieves textual content but also preserves the positional coordinates of text blocks, which are subsequently analyzed to detect multi-column structures and maintain logical reading order during parsing. In contrast, complex layouts with multi-column structures or embedded graphics are processed using IBM’s \href{https://www.docling.ai/}{\texttt{DocLing}} service, which provides layout-aware parsing capabilities. When the input consists of scanned or image-based documents, Optical Character Recognition (OCR) techniques are applied to recover textual content.

Following text extraction, the system performs several normalization steps to prepare the data for natural language processing. This includes removing unwanted symbols, correcting accented or malformed characters, standardizing whitespace and punctuation, and ensuring consistent text encoding. These preprocessing steps produce a clean and uniform text representation, minimizing downstream errors.

Once the text is standardized, the module proceeds to extract structured information across multiple categories, including personal details, education, professional experience, skills, and languages. To achieve this, \textbf{\textit{Smart-Hiring}} employs a hybrid approach that combines rule-based patterns, lightweight machine learning models, and heuristic logic.

\begin{itemize}
    \item \textbf{Names extraction:} Extracting candidate names is a particularly challenging task due to inconsistent formatting, and there may be partial loss of layout information during text extraction—especially for scanned or irregularly formatted CVs, which may still exhibit disrupted reading order. To address this, a supervised classifier was trained on a curated dataset of Algerian names obtained from Kaggle, achieving high accuracy in distinguishing personal names from surrounding text. The model accounts for common characteristics of Algerian names, such as multi-token family names, Arabic-origin spelling variations, and mixed use of Latin and accented characters.

    \item \textbf{Contact information:} Emails, phone numbers, and addresses are detected using regular expressions and heuristic rules within dedicated sections such as "Contact", "Profile",or "Personal Information". This strategy ensures robust extraction even in documents with noisy formatting or non-standard separators.

    \item \textbf{Skills extraction:} The system leverages a fuzzy matching algorithm against a comprehensive skills lexicon derived from LinkedIn. This allows it to identify both exact and near-miss skill mentions, capturing variations such as "JS" vs "JavaScript" or "data analysis" vs "data analytics".

    \item \textbf{Education level:} To determine the candidate’s highest education level, fuzzy matching is applied on a predefined list of academic degrees (e.g., Licence, Master, PhD, Engineer), handling both English and French variants.

    \item \textbf{Experience extraction:} The total years of experience are inferred from the "Experience" section by detecting and parsing date intervals, complemented by heuristic logic to estimate the duration of missing or partial entries.
\end{itemize}

The final output of this stage is a structured representation of each resume, capturing all relevant attributes in a unified schema. This structured data serves as the foundation for the subsequent job–candidate matching modules, enabling accurate, explainable, and efficient downstream processing.

\section{Job Matching}
The second stage of the \textbf{Smart-Hiring} pipeline is dedicated to the job matching process, which aims to identify the best-fit candidates for a given job description. This component operates by comparing the structured information extracted from resumes with requirements specified in job postings.

The input to this module consists of two main sources: the parsed resume generated by the information extraction component and the job description provided by recruiters. Both inputs contain complementary information that allows the system to evaluate candidate's job compatibility across multiple dimensions, including experience, education, and skills.

The core of the matching process lies the text representation and similarity computation mechanism. \textbf{Smart-Hiring} compares key profile attributes, such as the candidate’s address, years of experience, and educational level—against the corresponding requirements specified in the job description. For skill-based comparison, both resume and job description skills are encoded using the transformer-based sentence embedding model \texttt{all-MiniLM-L6-v2}. This model captures semantic similarity, allowing conceptually related skills—such as “Linux” and “Ubuntu”, to be recognized as compatible. Pairwise cosine similarity is then computed between the embedded skill vectors to quantify their semantic closeness. The \texttt{all-MiniLM-L6-v2} model was selected for its balance between computational efficiency and semantic accuracy. It effectively captures contextual similarity between multi-word skill phrases (e.g., “data analysis” vs “data analytics”) and generalizes across linguistic variations common in bilingual resumes.

Each matching criterion is assigned a specific weight reflecting its relative importance to the overall evaluation. Attribute weights are empirically determined based on domain expertise, ensuring that critical dimensions, such as core skills and experience, contribute more strongly to the overall matching score between a candidate’s resume and the job description. The final candidate's job compatibility score is obtained by aggregating these weighted similarities, producing a single numeric score that reflects the candidate’s overall suitability for the job.

To ensure transparency and trust in the matching results, the system integrates an explainability layer that highlights the most influential factors contributing to the final score. Matching keywords, overlapping skills, and aligned experience levels are visually emphasized, enabling recruiters to understand the reasoning behind each recommendation. This interpretability component supports informed decision-making and reduces bias by making the automated ranking process both clear and auditable.

The output of this module is a ranked list of candidates, each associated with a similarity score that represents their degree of alignment with the job description. This ranking facilitates rapid and data-driven candidate selection while maintaining interpretability and fairness throughout the hiring process.

\section{Experiments and Results Discussion}
To assess the performance and reliability of the \textbf{Smart-Hiring} system, several experiments were conducted using a collection of resumes and job descriptions obtained from various professional domains. The dataset included resumes and job descriptions from domains such as IT and telecommunications, written primarily in French and English. Although the dataset size was modest (around 1K CVs), it covered diverse formatting styles and linguistic variations representative of real-world recruitment scenarios.

The evaluation focused primarily on two aspects: the accuracy of resume information extraction and the effectiveness of the job matching module. The quality of extracted data was assessed through manual inspection and feedback from HR experts. For job matching, similarity results and ranked candidate lists were evaluated by HR teams to estimate the alignment between Smart-Hiring’s recommendations and human judgment. Accuracy and top-\textit{k} match rate metrics were used to quantify the system’s performance.

The information extraction module achieved high accuracy for structured entities such as education, contact details, and skills. The job–candidate matching module obtained a good top–3 match rate, indicating strong alignment with expert judgments. Overall, the pipeline produced consistent and interpretable results, with most of the top-ranked candidates corresponding closely to the expectations of human evaluators. These outcomes indicate that \textbf{Smart-Hiring} can significantly reduce manual workload during the pre-screening phase while maintaining transparency in its decision-making process.

However, some limitations were observed during the experiments. In particular, resumes with highly complex layouts posed challenges during the extraction phase. When the layout contained multiple columns, graphics, or dense formatting, the extracted text occasionally lost its original structure, which negatively impacted downstream entity recognition and matching accuracy. This issue highlights the importance of layout-preserving extraction methods in achieving robust performance.

The experiments also confirmed the value of incorporating explainability into the matching process. Unlike many existing systems that only compute embedding similarities between resumes and job descriptions, \textbf{Smart-Hiring} provides explicit reasoning behind its recommendations by highlighting matched skills and experience. This interpretability not only enhances recruiter confidence but also supports fairer and more accountable AI-assisted hiring.

These results highlight the potential of hybrid, explainable AI in recruitment contexts. By combining semantic embeddings with interpretable logic, \textbf{Smart-Hiring} bridges the gap between automation efficiency and human understanding—a key requirement for adoption in professional HR workflows.

Future improvements will focus on adopting more advanced and layout-aware text parsing techniques, which can better preserve document structure during extraction. Additionally, the integration of large language models (LLMs) may further enhance context understanding and semantic accuracy, leading to a more intelligent and explainable approach to information extraction and candidate-job matching.

\section{Conclusion} %%ok%%%
This paper presented \textbf{Smart-Hiring}, an end-to-end AI-based recruitment framework that integrates advanced Natural Language Processing techniques for both resume information extraction and Descriptions–job matching. Unlike conventional systems limited to keyword search or static rules, Smart-Hiring combines layout-aware document parsing, hybrid extraction strategies, and transformer-based semantic similarity modeling to deliver accurate, interpretable, and domain-agnostic recruitment insights.

Experimental evaluations on real-world datasets demonstrated that the proposed pipeline can effectively transform heterogeneous and unstructured resumes into standardized, analyzable data, enabling meaningful semantic comparison with job descriptions. The system’s explainability layer further enhances transparency by highlighting key factors driving candidate ranking decisions—offering recruiters not only efficiency but also trust and accountability in automated screening.

Beyond its technical performance, Smart-Hiring contributes as a practical framework that bridges the gap between unstructured text analytics and real-world HR applications. It illustrates how modern NLP can be responsibly deployed to reduce human workload while promoting fairness, interpretability, and reproducibility in data-driven hiring.

Future work will focus on expanding dataset diversity, incorporating multilingual support, and leveraging large language models (LLMs) to enhance contextual reasoning and bias detection. Moreover, integrating feedback loops between recruiters and the system which may enable continuous learning, leading toward adaptive, human–AI collaborative hiring. 

\bibliographystyle{unsrtnat}
\bibliography{references}  
\end{document}